\documentclass[sn-mathphys,Numbered]{sn-jnl}% Math and Physical Sciences Reference Style
\usepackage{graphicx}%
\usepackage{multirow}%
\usepackage{amsmath,amssymb,amsfonts}%
\usepackage{amsthm}%
\usepackage{mathrsfs}%
\usepackage[title]{appendix}%
\usepackage{xcolor}%
\usepackage{textcomp}%
\usepackage{manyfoot}%
\usepackage{booktabs}%
\usepackage{algorithm}%
\usepackage{algorithmicx}%
\usepackage{algpseudocode}%
\usepackage{listings}%
\usepackage{natbib}
\usepackage{doi}
\usepackage{booktabs}  % For better-looking tables
\usepackage{siunitx}   % For better number alignment
\usepackage{multirow}  % For multirow cells
\usepackage{caption}   % For customizing captions
\usepackage{amsmath}
\usepackage{hyperref}
\theoremstyle{thmstyleone}%
\theoremstyle{thmstyletwo}%

\theoremstyle{thmstylethree}%

\begin{document}

\title[Article Title]{Benchmarking PathCLIP for Pathology Image Analysis}


\author[1,2]{\fnm{Sunyi} \sur{Zheng}}%\email{zhengsunyi@westlake.edu.cn}
\equalcont{These authors contributed equally to this work.}

\author[1]{\fnm{Xiaonan} \sur{Cui}}%\email{cuixiaonan1988@163.com}
\equalcont{These authors contributed equally to this work.}

\author[3]{\fnm{Yuxuan} \sur{Sun}}%\email{chenpingyi@westlake.edu.cn}
\author[3]{\fnm{Jingxiong} \sur{Li}}%\email{sunyuxuan@westlake.edu.cn}
\author[3]{\fnm{Honglin} \sur{Li}}%\email{lijingxiong@westlake.edu.cn}
\author[3]{\fnm{Yunlong} \sur{Zhang}}%\email{lihonglin@westlake.edu.cn}
\author[3]{\fnm{Pingyi} \sur{Chen}}%\email{zhangyunlong@westlake.edu.cn}
\author[4]{\fnm{Xueping} \sur{Jing}}%\email{x.jing@umcg.nl}
\author[1]{\fnm{Zhaoxiang} \sur{Ye}}%\email{zye@tmu.edu.cn}
\author*[2]{\fnm{Lin} \sur{Yang}}\email{yanglin@westlake.edu.cn}

%\orgdiv{Key Laboratory of Cancer Prevention and Therapy, Department of Radiology}, 
\affil[1]{\orgname{Tianjin Medical University Cancer Institute and Hospital, National Clinical Research Center for Cancer, Tianjin’s Clinical Research Center for Cancer, Key Laboratory of Cancer Prevention and Therapy, Department of Radiology}, \orgaddress{ \city{Tianjin}, \country{China}}}

\affil[2]{\orgdiv{School of Engineering}, \orgname{Westlake University}, \orgaddress{ \city{Hangzhou}, \country{China}}}

%\orgdiv{College of Computer Science and Technology}, 
\affil[3]{\orgname{Zhejiang University}, \orgaddress{ \city{Hangzhou}, \country{China}}}

\affil[4]{\orgdiv{Department of Radiation Oncology}, \orgname{University Medical Center of Groningen}, \orgaddress{ \city{Groningen}, \country{The Netherlands}}}
% \affil[3]{\orgdiv{Department}, \orgname{Organization}, \orgaddress{\street{Street}, \city{City}, \postcode{610101}, \state{State}, \country{Country}}}


\abstract{Accurate image classification and retrieval are of importance for clinical diagnosis and treatment decision-making. The recent contrastive language-image pretraining (CLIP) model has shown remarkable proficiency in understanding natural images. Drawing inspiration from CLIP, PathCLIP is specifically designed for pathology image analysis, utilizing over 200,000 image and text pairs in training. While the performance the PathCLIP is impressive, its robustness under a wide range of image corruptions remains unknown. Therefore, we conduct an extensive evaluation to analyze the performance of PathCLIP on various corrupted images from the datasets of Osteosarcoma and WSSS4LUAD. In our experiments, we introduce seven corruption types including brightness, contrast, Gaussian blur, resolution, saturation, hue, and markup at four severity levels. Through experiments, we find that PathCLIP is relatively robustness to image corruptions and surpasses OpenAI-CLIP and PLIP in zero-shot classification. Among the seven corruptions, blur and resolution can cause server performance degradation of the PathCLIP. This indicates that ensuring the quality of images is crucial before conducting a clinical test. Additionally, we assess the robustness of PathCLIP in the task of image-image retrieval, revealing that PathCLIP performs less effectively than PLIP on Osteosarcoma but performs better on WSSS4LUAD under diverse corruptions. Overall, PathCLIP presents impressive zero-shot classification and retrieval performance for pathology images, but appropriate care needs to be taken when using it. We hope this study provides a qualitative impression of PathCLIP and helps understand its differences from other CLIP models.}




\maketitle

\section{Introduction}
In recent years, with the digitization of pathology slides, artificial intelligence (AI) has rapidly integrated into the diagnostic process, leading to a significant transformation in clinical pathology \cite{campanella2019clinical,chen2021annotation,fremond2023interpretable}. The synergy between digital pathology and AI has paved the way for automating and redefining diagnostic procedures. This includes accurate cell recognition \cite{wang2022deep,shui2023deformable}, cancer region segmentation \cite{saltz2018spatial,li2020deep}, image retrieval for diagnosis \cite{wang2023retccl,huang2023visual,sun2023pathasst}, and the identification of cancer subtypes \cite{woerl2020deep,li2023task, cui2023prediction}. As these AI systems continue to evolve, they not only promise to streamline the diagnostic workflow but also enhance the accuracy and efficiency of pathology analyses. Ultimately, these advancements are expected to provide valuable support to healthcare professionals.

The artificial intelligence landscape has undergone a revolution with the emergence of large-scale language models. Leveraging transformer architectures, these language models possess the ability to respond to free-text queries without specific training for the given task. An example is the large language model meta AI (LLaMA) \cite{touvron2023llama}, which offers various model versions with parameters ranging from 7 billion to 65 billion. All these versions are trained using masked language modeling and next-word prediction techniques with training data sourced from publicly available repositories like Wikipedia, Common Crawl. The results of LLaMA demonstrate that cutting-edge performance in tasks such as reading comprehension and code generation can be attained without reliance on proprietary datasets. ChatGPT \cite{radford2018improving} also plays a pivotal role in this advancement. Trained on a vast corpus of text data extracted from books, articles and web pages, ChatGPT has developed a profound understanding of the intricacies and nuances of natural language. One of its remarkable abilities is the generation of text that closely mimics human language when prompted, making it valuable for a range of natural language processing tasks, including chatbots, language translation, and text summarization. Another promising foundation approach is named ChestXRayBERT \cite{cai2021chestxraybert}. This approach utilizes a pre-trained BERT-based language model \cite{devlin2018bert} to automatically generate the impression section of chest radiology reports. This approach has the potential to significantly reduce the workload of radiologists and enhance communication between radiologists and referring physicians. In experiments, ChestXRayBERT outperforms existing state-of-the-art models in terms of readability, factual correctness, informativeness and redundancy.

Apart from focusing on large-scale language models, researchers are also dedicating attention to expansive multi-modal models within the field of computer vision. An example is the segment anything model \cite{kirillov2023segment}. It is a segmentation system capable of zero-shot generalization to unfamiliar objects and images without requiring additional training. It employs a ViT-H image encoder \cite{dosovitskiy2020image} for image embedding and a prompt encoder to produce prompt embeddings. After encoding, its mask decoder which is based on a lightweight transformer predicts object masks using the image and prompt embeddings. Another recently introduced multi-modal model is VisualGPT \cite{chen2022visualgpt}, which combines a pre-trained language model of GPT-2 \cite{radford2019language} and a vision model of ResNet-101. It incorporates an encoder-decoder attention mechanism with an unsaturated rectified gating function to bridge the semantic gap between different modalities. VisualGPT has exhibited state-of-the-art performance on a medical report generation dataset of IU X-ray and has surpassed strong baseline models on the MS COCO data. Furthermore, GPT-4 with vision (GPT-4V) \cite{yang2023dawn} also showcases promising capabilities in analyzing text, images, and voice. Different from GPT-2, GPT-4v allows users to instruct GPT-4 for analyzing user-provided image inputs. In the context of aided medical diagnosis \cite{yan2023multimodal}, GPT-4V can provide cautious responses and generate accurate localization if appropriate cues are provided. The utilization of these large models marks a significant advancement at the intersection of technology and healthcare, holding tremendous promise for more precise and nuanced image analysis.

To create a robust large multi-modal model, a robust vision model is crucial. One of widely used vision models is the contrastive language-image pre-training model (CLIP) \cite{radford2021learning}. The objective of the CLIP is to maximize similarity among positive samples while minimizing similarity among negative samples, facilitating the development of meaningful visual-semantic representations. What sets CLIP apart is its ability to comprehend a wide range of image-text pairs without explicit supervision, enabling excellence in various visual and language tasks, such as zero-shot image classification and image retrieval. Inspired by the realization that modern pre-training methods can benefit from aggregate supervision in web-scale text collections, OpenAI utilizes web data instead of crowd-labeled datasets such as ImageNet to create OpenAI-CLIP. The zero-shot performance of OpenAI-CLIP proves more resilient to distribution shifts than standard ImageNet models. However, since OpenAI-CLIP is predominantly trained on natural images, its performance may be suboptimal when applied to medical data. To address this issue for medical tasks, especially in pathological data analysis, pathology language–image pre-training (PILP) \cite{huang2023visual} and PathCLIP \cite{sun2023pathasst} are proposed. PILP is trained on data from Twitter and the LAION dataset \cite{schuhmann2022laion}, which contains pathology data from various internet sources. In contrast, PathCLIP leverages data from PubMed, books, and the WebPathology pathology atlas website, achieving state-of-the-art performance in pathology image analysis among CLIP series models. However, unlike CLIP, which has been evaluated on computer vision datasets, the robustness of PathCLIP has not been systematically benchmarked.

To bridge this gap, we have undertaken the present study, and our contributions can be summarized as follows:

• We investigate the robustness of PathCLIP across various corruptions and tasks on pathology datasets, providing insights into the challenges it might face when deployed in the real world.

• Our experiments on image corruptions indicate that the PathCLIP is somewhat robust to corruptions. But still blur and resolution can significantly affect the performance of the PathCLIP. Therefore, it is important to ensure image quality before using PathCLIP.

• Exploring different pathological tasks, we find that PathCLIP can achieve better performance than PLIP and OpenAI-CLIP in zero-shot classification. However, PLIP can surpass PathCLIP in image-image retrieval. It is advisable to selectively employ a CLIP model based on the specific tasks during pathology image analysis.

The rest of this article is organized as follows: Section 2 describes the related work, specifically benchmarking CLIP on natural images and deep learning algorithms on pathology images. Section 3 provides details on datasets and the generation of corrupted images, followed by Section 4 presenting experimental results and related analysis. Section 5 outlines the summary of the paper and discusses future work.

\section{Related work}
\subsection{Robustness assessment of CLIP on natural images}
Among these large multi-modal models, CLIP has caused a tremendous stir in natural image analysis, prompting many to conduct validation studies on natural images. The capacity of CLIP to diminish the necessity for task-specific training data opens doors for automating various specialized tasks. The model allows users to define image classification classes using natural language. However, this might introduce the potential to influence bias manifestation when applying CLIP in real situations. The investigation by Agarwal et al. \cite{agarwal2021evaluating} shows that CLIP may inherit biases from previous computer vision systems, raising concerns about ensuring safe behavior in its diverse and unpredictable applications. They find that 16.5\% of male images are misclassified into classes related to crime. These findings contribute to the growing call for redefining a safer and more trustworthy model, emphasizing considerations beyond task-oriented accuracy in the evaluation of model deployment. To measure the vulnerability of CLIP to frequency perturbations, Galindo et al. \cite{galindounderstanding} perform image generation and inpainting tasks for assessing robust features. In the experiments, the CLIP model presents lower robustness to lower frequency perturbations, indicating a higher dependence on features with lower frequency. The study conducted by Radford et al. \cite{radford2021learning} evaluates CLIP using a linear probe. Experimental findings reveal that the transfer scores of linear probes when trained on CLIP model representations, surpass those of alternative models with equivalent performance on ImageNet. This indicates CLIP is more robust to task shift in contrast to models that undergo pre-training on ImageNet. The insights derived from the aforementioned studies offer valuable evidence for individuals whose expertise lies outside the realms of AI or computer science. These findings can help them envision the potential applications of CLIP and may contribute to improving proficiency across various professional domains.

\subsection{Corruption analysis on pathology images}
While deep learning models have demonstrated remarkable results in medical image processing \cite{zheng2022chrsnet, jing2023localization, zheng2023survival}, there is limited research on the stability of deep learning models in various situations. To assess the applicability of deep learning across different pathology datasets, Zhang et al. \cite{zhang2022benchmarking} conduct a benchmark study on three common convolutional neural networks and transformers. The evaluation is performed on lymph node sections from breast cancer metastases and images of cervical cancer. The results indicate a substantial decrease in accuracy and unreliable confidence estimation of deep learning models when confronted with corrupted images. Similarly, Zhang et al. \cite{zhang2020corruption} investigate comprehensive corruption types, including bubble, shadow, color cast, exposure, defocus and stitching for peripheral blood smears. They evaluate ResNet and DenseNet to explore the effect of image corruption on model performance. Results suggested that deep learning models are sensitive to color cast in blood cell images. Additionally, Huang et al. \cite{huang2023assessing} analyze the physical causes of full-stack corruptions throughout the pathological life cycle. They propose an omni-corruption emulation method to reproduce corruptions. The study finds that using corrupted datasets as augmentation data for training and experiments can enhance the generalization ability of the models. It is noteworthy that the aforementioned papers mainly focus on non-foundation models. With the growing importance of large foundation models, it is necessary to have a comprehensive understanding of them before using these foundation models.

\section{Methods}
In this section, we evaluate and compare the robustness of CLIP models for the tasks of image classification and image-image retrieval on pathology images related to bone cancer and lung cancer. In the following sections, we provide data descriptions, types of image corruptions, evaluation metrics, and implementation details.

\subsection{Datasets}
\textbf{Osteosarcoma}: AA clinical scientist team at the university of Texas southwestern medical center collected this dataset in order to support the development of AI for diagnosing Osteosarcoma in adolescent patients with bone cancer \cite{mishra2017histopathological, leavey2019osteosarcoma}. The dataset comprises hematoxylin and eosin (H\&E) stained osteosarcoma histology images sourced from archival samples of 50 patients treated at Children's Medical Center, Dallas, between 1995 and 2015. The set consists of 1144 images, each possessing dimensions of 1024 x 1024 pixels at 10X resolution. These images are categorized into three groups of non-tumor, viable tumor, and necrosis, depending on the prevalent cancer type. Two medical professionals conduct the annotation process, wherein all images are distributed between two pathologists. Each image received a singular annotation, with a specific pathologist responsible for annotating any given image. As a result, we include 536 (47\%) non-tumor tiles, 263 (23\%) necrotic tumor tiles, and 345 (30\%) viable tumor tiles for analysis.

\textbf{WSSS4LUAD}: The WSSS4LUAD challenge involves 67 H\&E stained slides from Guangdong provincial people's hospital and 20 Whole Slide Images from the cancer genome atlas\cite{han2022multi}. The primary objective of this challenge is to achieve pixel-level prediction of tissue types, thereby significantly reducing annotation efforts. In the challenge, participants are provided with image-level annotations for machine learning algorithm training and pixel-level ground truth for validation and testing. Because image-level annotations are exclusively available in the training set, we utilize this set for evaluating the performance of models. Lastly, we take 6574 pure tumor patches and 1832 completely normal patches from the training set.

\subsection{Image Corruptions}
Evaluating model robustness in the face of image corruptions can involve examining different aspects such as common image corruptions \cite{zhang2022benchmarking}, style transfer \cite{qiao2023robustness} and adversarial attacks \cite{zhang2023attack}. Although the latter two methods are useful for assessing neural network performance, they are relatively uncommon in real-world scenarios. Hence, our primary focus is on seven commonly encountered image corruptions, which occur either during the creation of slices due to variations in the proportions of staining reagents or are influenced by device parameters in the scanning process. These corruptions encompass brightness, contrast, Gaussian blur, resolution, saturation, hue, and markup, as visually depicted in Figure 1. 

\begin{figure*}[!tb]
\centering
\includegraphics[scale=0.5]{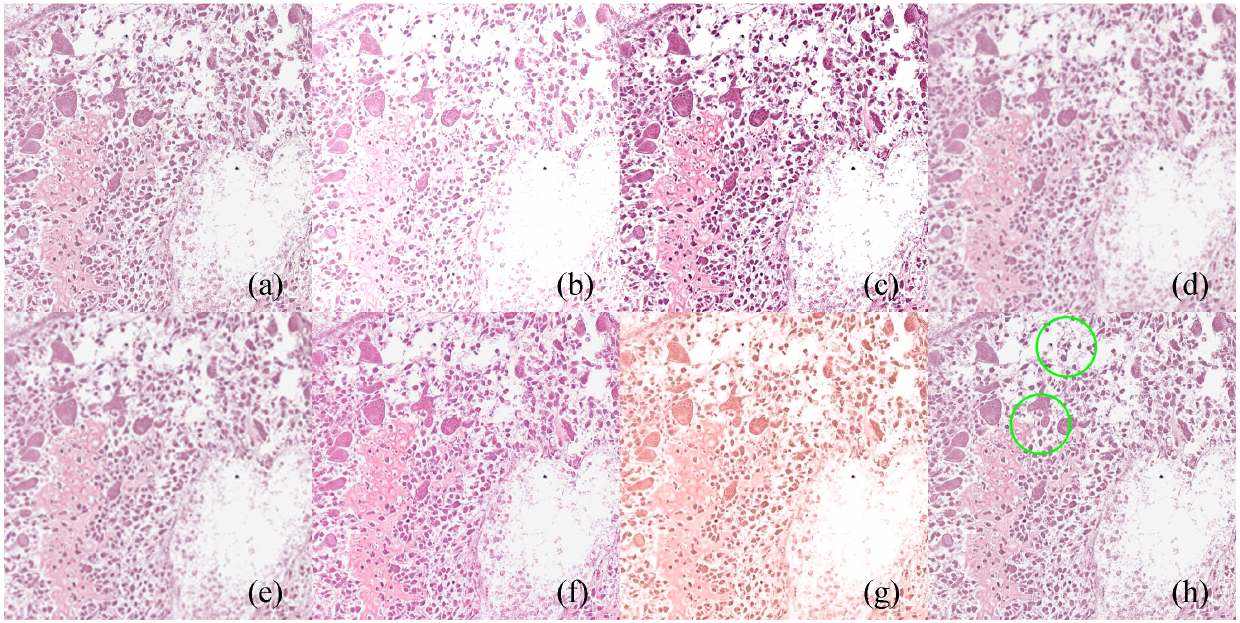}
\caption{\label{fig1} Examples of corrupted images. (a)-(h) represent the corruption of brightness, contrast, Gaussian blur, resolution, saturation, hue, and markup, respectively.}
\end{figure*}

Specifically, we make modifications to the appearance in terms of brightness and contrast to replicate high exposure and contrast enhancement. We also utilize Gaussian blur to simulate the effects of a microscope being out of focus, while variations in resolution are employed to mimic data from images that have been either magnified or reduced in size. Besides, elements such as saturation and hue, are also incorporated into our simulated image corruptions to introduce diverse color styles. Markup is considered to emulate the situation when pathologists annotate images to highlight certain tumor areas. For above corruptions, each type has four severity levels, manifesting at different intensities. All corruption types are implemented as functions, allowing their application to data during testing while saving storage space. In total, we apply 28 corruptions to the data, taking into account the anticipated diversity in corruption types and their intensities as observed in real-world scenarios.

\subsection{CLIP framework}
The contrastive language-image pre-training (CLIP) model, introduced by OpenAI, represents an advanced approach to multimodal learning. It aims to enhance visual and semantic comprehension by simultaneously learning from language and images using contrastive learning theory. The model seeks to maximize the similarity of positive samples while minimizing the similarity of negative samples, enabling the acquisition of meaningful visual-semantic representations. What sets CLIP apart is its ability to comprehend a broad spectrum of image-text pairs without requiring explicit supervision. This universality allows CLIP to excel in various visual and language tasks, including zero-shot image classification and image retrieval.

The architecture of the CLIP model comprises two key components of an image encoder and a text encoder. The Image Encoder employs either a convolutional neural network architecture or a transformer to extract high-level features from images, while the text encoder adopts a transformer architecture, focusing on converting text into semantic vector representations.

\subsection{Evaluation metric}
To quantitatively assess the impact of various corruptions on CLIP models, we employ metrics including accuracy and \(F1\) score for the zero-shot classification task, and precision for the image retrieval task.

Specifically, to ensure an equitable comparison across models on data with diverse distributions in zero-shot classification, results are presented for accuracy and \(F1\) score. Accuracy, measuring the percentage of correct predictions in a classification model, is well-suited for balanced class distributions. It is computed as the ratio of true positive and true negative predictions to the total instances. The formula of accuracy is shown below:

\begin{equation}
accuracy=\frac{TP + TN}{TP + TN + FP +FN}
\end{equation}

where \(TP\), \(FP\), \(TN\), and \(FN\) represent true positive, false positive, true negative, and false negative cases, respectively.

On the other hand, the F1 score is particularly valuable in scenarios with imbalanced class distributions, where one class significantly outnumbers the others. It is a metric that provides a balanced assessment of model performance by considering both precision and recall. This unified measure addresses false positives and false negatives. The function of the \(F1\) score is as follows:

\begin{equation}
F_1=\frac{2TP}{2TP + FP +FN}
\end{equation}

Similar to zero-shot classification, which can identify the closest text from a pool of candidates given an image, image-image retrieval is a technique that can identify the closest image from a pool of candidates given an image. This is achieved by directly calculating the cosine similarity of each paired image-image under the same embedding space. The performance of image-image retrieval is evaluated by class retrieval precision across models. We consider the retrieved results correct if the class of the given image matches the class of the top \(K\) retrieved images. This means the given image successfully hits all (HA) the retrieved ones. The function of the precision, \(HA@K\), is defined as:

\begin{equation}
I(A) = \begin{cases} 
    1, & \text{if } A \text{ satisfies the condition} \\
    0, & \text{otherwise}
\end{cases}
\end{equation}

\begin{equation}
HA@K = \frac{1}{n} \sum_{i=1}^{n} I\left(\forall\{ R_1, R_2, \ldots, R_k \} = G_i\right)
\end{equation}

where \(n\) and \(k\) are the numbers of images in the test set and the top \(k\) retrieved images, respectively. \(\{ R_k \}\) represents the class of the retrieved image \(k\), whereas \(\{ G_i \}\) is the class of the given image. \(I(A)\) is the function to check if the given images share the same class as the top \(K\) retrieved images.

\subsection{Implementation details}
All experiments are performed on an NVIDIA GPU of V100 using PyTorch. The image size of the model input is 224x224. The top \(k\) in the image-image retrieval has two settings of 5 and 10. The key corruption parameters for brightness, contrast, saturation, and hue are set at 0.4, 0.8, 1.2, and 1.6 for four severity levels, respectively. For blur and markup, the parameter values are 1, 2, 3, and 4, respectively. In the resolution setting, we decrease the image size by 20\%, 40\%, 60\%, and 80\% from the severity level 1 to 4. Severity levels ranging from mild to severe are denoted as S-A to S-D.

\section{Experiments and results}
\subsection{Performance on zero-shot image classification}
To comprehensively investigate the robustness of PathCLIP, we use two hispathological datasets related to bone cancer and lung cancer, subjecting them to seven common corruptions at four severity levels. The performance the model in zero-shot classification is presented in Table 1.

\begin{table}[!ht]
    \centering
    \caption{Performance of PathCLIP for zero-shot classification on two pathology datasets. The best results are highlighted in bold.}
    \label{t1}
    \begin{tabular}{lcccccc}
    \toprule
        Corruption & Severity level & \multicolumn{2}{c}{Osteosarcoma} & \multicolumn{2}{c}{WSSS4LUAD} \\
    \cmidrule(lr){3-4} \cmidrule(lr){5-6}
        & & {Accuracy} & {F1 score} & {Accuracy} & {F1 score} \\
    \midrule
        \multirow{1}{*}{Origin} & 0 & 0.697  & 0.664  & 0.921  & 0.923  \\
    \midrule
        \multirow{4}{*}{Brightness} 
        & -2 & 0.605  & 0.527  & 0.653  & 0.682  \\
        & -1 & \textbf{0.637}  & \textbf{0.584}  & 0.763  & 0.783  \\
        & 1 & 0.632  & 0.582  & 0.770  & 0.789  \\
        & 2 & 0.594  & 0.535  & \textbf{0.829}  & \textbf{0.841}  \\
    \midrule
        \multirow{4}{*}{Contrast} 
        & -2 & 0.582  & 0.496  & 0.637  & 0.667  \\
        & -1 & 0.634  & 0.583  & 0.744  & 0.766  \\
        & 1 & 0.631  & 0.577  & 0.720  & 0.745  \\
        & 2 & \textbf{0.635}  & \textbf{0.584}  & \textbf{0.752}  & \textbf{0.773}  \\
    \midrule
        \multirow{4}{*}{Blur} 
        & 1 & \textbf{0.635}  & \textbf{0.582}  & \textbf{0.524}  & \textbf{0.546}  \\
        & 2 & 0.616  & 0.550  & 0.256  & 0.158  \\
        & 3 & 0.603  & 0.521  & 0.222  & 0.087  \\
        & 4 & 0.595  & 0.500  & 0.218  & 0.078  \\
    \midrule
        \multirow{4}{*}{Resolution} 
        & 1 & \textbf{0.635}  & \textbf{0.584}  & \textbf{0.715}  & \textbf{0.740}  \\
        & 2 & 0.634  & 0.582  & 0.618  & 0.647  \\
        & 3 & 0.632  & 0.579  & 0.421  & 0.420  \\
        & 4 & 0.600  & 0.523  & 0.224  & 0.091  \\
    \midrule
        \multirow{4}{*}{Saturation} 
        & -2 & 0.593  & 0.504  & 0.718  & 0.743  \\
        & -1 & 0.628  & 0.571  & \textbf{0.732}  & \textbf{0.756}  \\
        & 1 & 0.636  & 0.587  & 0.720  & 0.744  \\
        & 2 & \textbf{0.641}  & 0.\textbf{595}  & 0.710  & 0.735  \\
    \midrule
        \multirow{4}{*}{Hue} 
        & -2 & 0.564  & 0.465  & 0.385  & 0.366  \\
        & -1 & 0.584  & 0.494  & \textbf{0.794}  & \textbf{0.810}  \\
        & 1 & \textbf{0.652}  & \textbf{0.603}  & 0.747  & 0.769  \\
        & 2 & 0.594  & 0.517  & 0.622  & 0.650  \\
    \midrule
        \multirow{4}{*}{Markup} 
        & 1 & \textbf{0.573}  & \textbf{0.499}  & \textbf{0.503}  & \textbf{0.521}  \\
        & 2 & 0.570  & 0.493  & 0.460  & 0.468  \\
        & 3 & 0.570  & 0.489  & 0.445  & 0.449  \\
        & 4 & 0.563  & 0.478  & 0.435  & 0.436 \\
    \bottomrule
    \end{tabular}
\end{table}

Compared to the results on original images, the model exhibits varying levels of performance degradation in both accuracy and \(F1\) score when tested on corrupted images within the Osteosarcoma and WSSS4LUAD datasets, as illustrated in Table 1. With the increase of the severity level, there is a trend of performance degradation in images affected by blur and resolution in WSSS4LUAD. This is reasonable since more severe blur diminishes image clarity leading to the loss of pathology tissue structure or cell morphological features. Similarly, lower image resolution hinders the preservation of intricate details, rendering it less conducive to the meticulous analysis of pathological images by the model. The image color factor contributing to performance degradation is hue. PathCLIP has relatively stable results in terms of saturation and contrast regardless of the change of severity levels. Additionally, model performance can decrease from 0.923 to 0.682 if the severity level of brightness is set to -2. This emphasizes the importance of ensuring consistency between the parameters used in device settings and those used in training the model to enhance model robustness in practical applications. Regarding markup, an increased number of delineations leads to more information being obscured in pathology images, preventing them from providing complete feature information and causing a decline in model performance. This deduction aligns with the outcomes derived from our experimental investigations.

\begin{figure*}[!tb]
\centering
\includegraphics[scale=0.42]{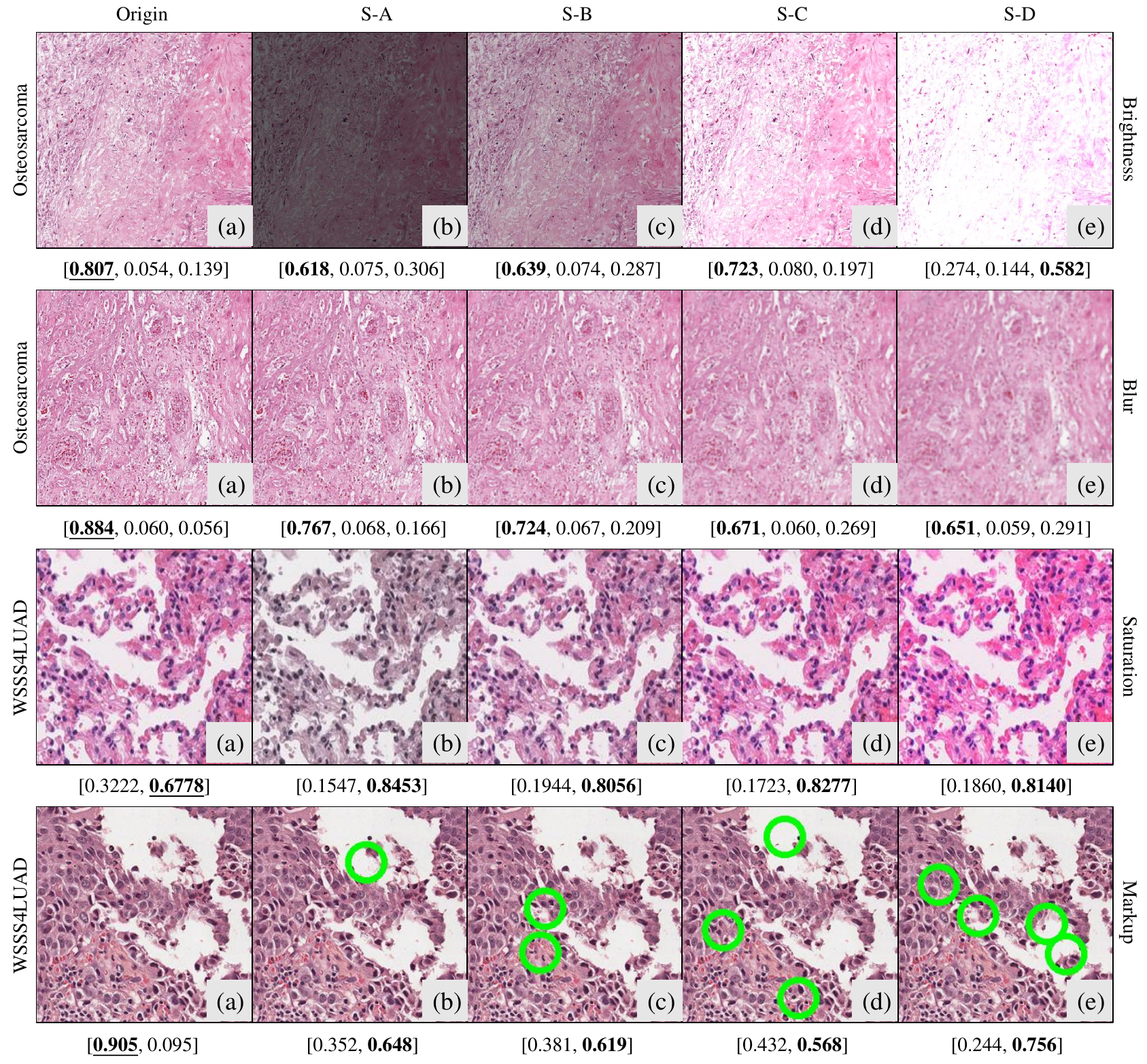} 
\caption{\label{fig2} Performance of PathCLIP on image corruptions varies across four severity levels, ranging from mild to severe, denoted as S-A to S-D. The bold values represent the maximum prediction probabilities, while the underlined values indicate the ground truth classes.}
\end{figure*}

To further explore the impact of corruptions on model performance, we analyze the detailed prediction results of images. In Figure 2, we present examples illustrating how prediction outcomes change after applying various corruptions to images. In the first row of results, we observe that altering the brightness of the image, while maintaining the correct predicted category, led to a decrease in the confidence of the model. When the image brightness is excessively high, the image appears overly white, resembling a background image. Therefore, the model misclassifies it as an image without tumors. In the case of blur, we find that increased blurriness results in a lower probability of the model predicting the presence of necrotic tumors. Saturation is a common image enhancement technique, and many studies have used it in model development \cite{tellez2019quantifying, takahashi2019data}. In the results of saturation, we can see that after enhancing the image, the confidence of the model in predicting the category increases. This enhancement may be attributed to saturation emphasizing features in the image containing tumor epithelial tissue. Besides, it is interesting that the model initially correctly predicted the image category. But after adding a markup, it fails to perform correct prediction. Furthermore, when the markup reached severity level 4, the model tends to exhibit higher confidence in incorrect category predictions. This phenomenon can be explained by the situation that the limited exposure of the model to an insufficient number of markups during the training process. This limitation cause the model being excessively confident when predicting categories for images with previously unseen markups.

\subsection{Performance on image-image retrieval}
In clinical practice, the swift and accurate retrieval of relevant medical images can potentially facilitate quicker decision-making and improve patient outcomes. Therefore, we also evaluate the performance of the model in image-image retrieval under various image corruptions. The detailed results of PathCLIP are depicted in Table 2.

\begin{table}[!ht]
    \centering
    \caption{Performance of PathCLIP for image-image retrieval on two pathology datasets. Best results are highlighted in bold.}
    \label{t2}
    \begin{tabular}{lcccccccc}
    \toprule
        Corruption & Severity & \multicolumn{2}{c}{Osteosarcoma} & \multicolumn{2}{c}{WSSS4LUAD} \\
    \cmidrule(lr){3-4} \cmidrule(lr){5-6}
        & Level & HA@5 & HA@10 & HA@5 & HA@10 \\
    \midrule
        \multirow{1}{*}{Origin} & 0 & 0.785 & 0.684 & 0.948 & 0.915 \\
    \midrule
        \multirow{4}{*}{Brightness}
        & -2 & 0.732 & 0.609 & 0.916 & 0.864 \\
        & -1 & 0.744 & 0.619 & 0.920 & 0.868 \\
        & 1 & \textbf{0.777} & \textbf{0.652} & \textbf{0.930} & \textbf{0.884} \\
        & 2 & 0.729 & 0.624 & 0.926 & 0.878 \\
    \midrule
        \multirow{4}{*}{Contrast} 
        & -2 & 0.736 & 0.619 & 0.921 & \textbf{0.874} \\
        & -1 & 0.743 & 0.628 & 0.921 & 0.870 \\
        & 1 & \textbf{0.755} & 0.643 & 0.922 & 0.869 \\
        & 2 & 0.743 & \textbf{0.651} & \textbf{0.924} & 0.871 \\
    \midrule
        \multirow{4}{*}{Blur} 
        & 1 & \textbf{0.747} & 0.631 & \textbf{0.911} & \textbf{0.857} \\
        & 2 & 0.744 & \textbf{0.641} & 0.830 & 0.733 \\
        & 3 & 0.739 & 0.614 & 0.772 & 0.652 \\
        & 4 & 0.731 & 0.616 & 0.744 & 0.618 \\
    \midrule
        \multirow{4}{*}{Resolution} 
        & 1 & \textbf{0.752} & 0.636 & \textbf{0.914} & \textbf{0.861} \\
        & 2 & 0.750 & 0.637 & 0.910 & 0.857 \\
        & 3 & 0.749 & 0.640 & 0.879 & 0.810 \\
        & 4 & 0.749 & \textbf{0.646} & 0.768 & 0.644 \\
    \midrule
        \multirow{4}{*}{Saturation} 
        & -2 & 0.742 & 0.613 & 0.922 & 0.874 \\
        & -1 & 0.753 & 0.628 & 0.922 & 0.871 \\
        & 1 & 0.748 & 0.640 & 0.922 & 0.870 \\
        & 2 & \textbf{0.757} & \textbf{0.645} & \textbf{0.925} & \textbf{0.875} \\
    \midrule
        \multirow{4}{*}{Hue} 
        & -2 & 0.733 & 0.609 & 0.917 & 0.873 \\
        & -1 & \textbf{0.764} & 0.635 & \textbf{0.933} & \textbf{0.895} \\
        & 1 & 0.763 & 0.633 & 0.923 & 0.876 \\
        & 2 & 0.762 & \textbf{0.654} & 0.911 & 0.867 \\
    \midrule
        \multirow{4}{*}{Markup} 
        & 1 & 0.722 & \textbf{0.603} & \textbf{0.901} & \textbf{0.842} \\
        & 2 & \textbf{0.736} & 0.597 & 0.892 & 0.828 \\
        & 3 & 0.706 & 0.587 & 0.889 & 0.829 \\
        & 4 & 0.725 & 0.602 & 0.887 & 0.823 \\
    \bottomrule
    \end{tabular}
\end{table}

In the context of image-image retrieval, the performance of PathCLIP exhibits varying degrees of degradation across 7 corrupted images. When applying markup to test images, the model achieves its poorest performance compared to the use of original images. This again may indicate the potentially limited inclusion of markup data during the training process of PathCLIP. Furthermore, alterations in image severity levels for brightness, contrast, saturation, and hue do not substantially impact the performance of the model. This may suggest that PathCLIP is relatively robust to these effects brought about by color or light in image-image retrieval in image-image retrieval. Conversely, severe corruptions in blur and resolution reduce model performance on the WSSS4LUAD dataset, particularly concerning the HA@10 metric. Hence, it is essential to ensure image clarity to facilitate effective image retrieval in practical applications of the model.

\begin{figure*}[!tb]
\centering
\includegraphics[scale=0.38]{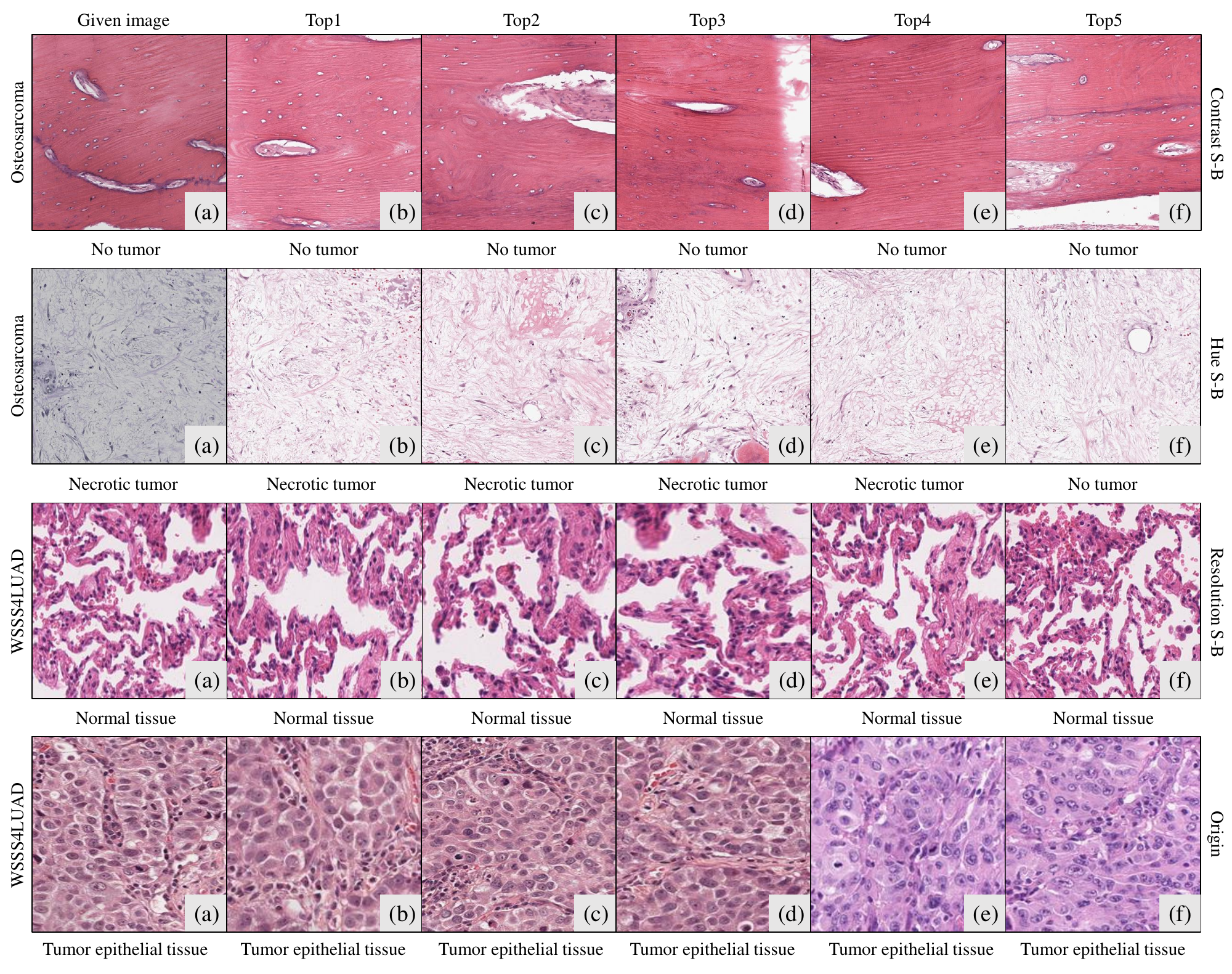}
\caption{\label{fig3} Example results of PathCLIP on image corruptions with the correct class name displayed below each image.}
\end{figure*}

We also examine the retrieved image results, and examples are illustrated in Figure 3. From the figure, it can be observed that the top 5 retrieved images exhibit a high degree of similarity to the provided images in terms of tissue structure and image features. And their image categories are nearly identical. However, in retrieving necrotic tumor images, one identified image lacks a tumor, despite its style being similar to that of the given image. This piece of evidence show that enhancing the performance of the model may require additional training data related to bone cancer. Furthermore, the last row of data reveals that two images with different background colors are also retrieved. This implies that the model on the WSSS4LUAD dataset may rely on analyzing image features such as pathological tissue structure and cells, rather than solely depending on background color.

\begin{figure*}[!tb]
\centering
\includegraphics[scale=0.45]{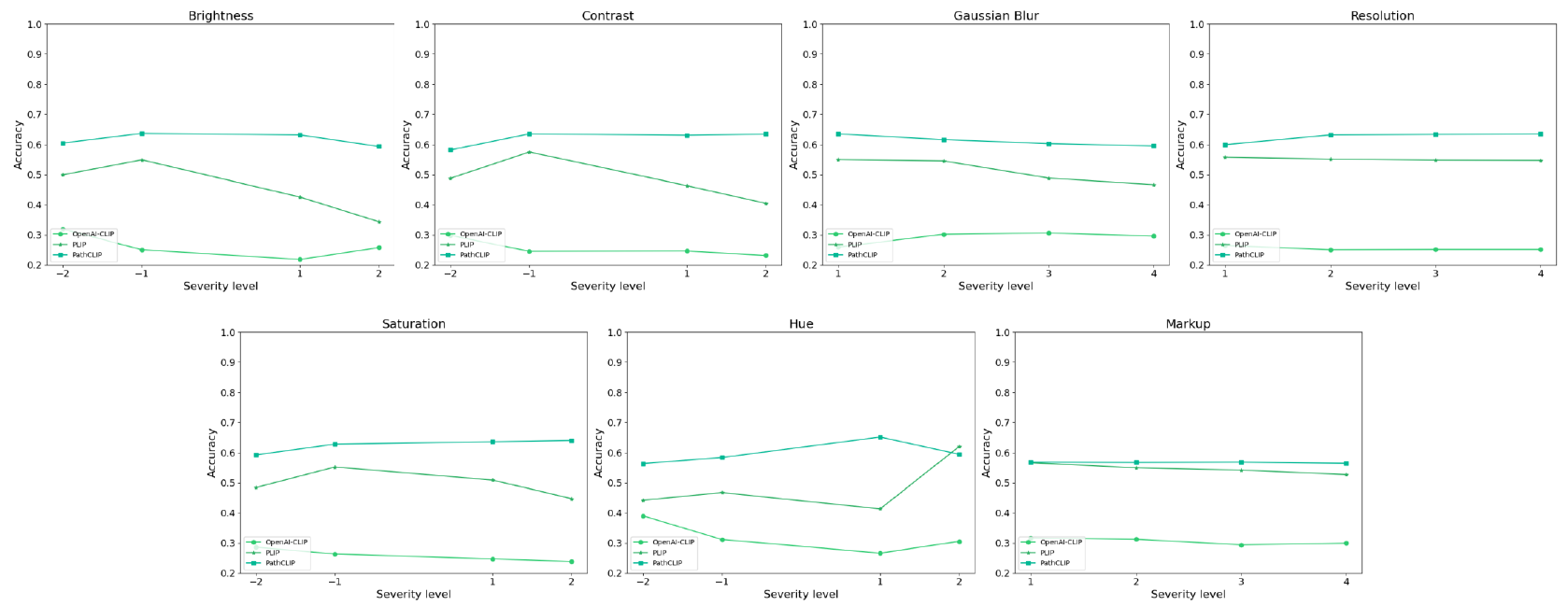}
\caption{\label{fig4} Model comparisons on zero-shot classification.}
\end{figure*}

\begin{figure*}[!tb]
\centering
\includegraphics[scale=0.45]{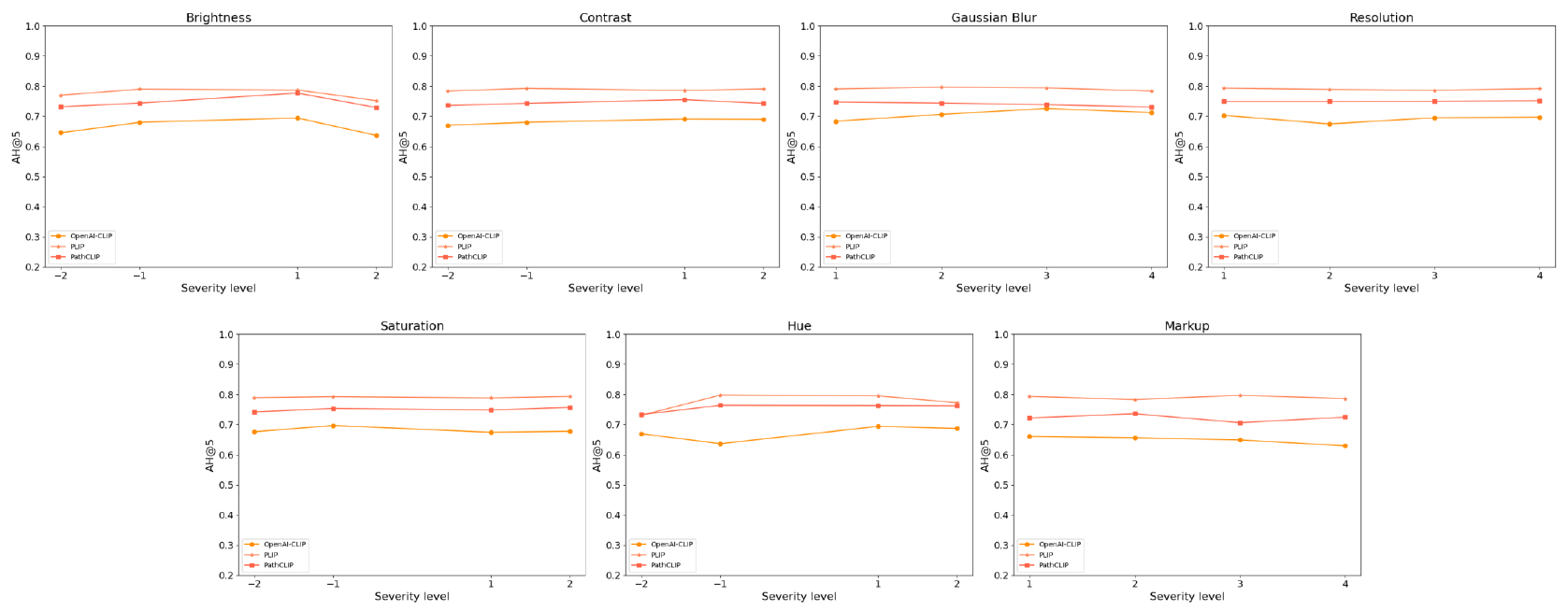}
\caption{\label{fig6} Model comparisons on image-image retrieval.}
\end{figure*}

\subsection{Model comparisons}
Since the Osteosarcoma dataset has three classes, it is more challenging for models to produce correct results than the WSSS4LUAD dataset which have two classes. We utilize Osteosarcoma for model comparisons and results are presented in figures 4 and 5.

PathCLIP achieves the best results in the zero-shot classification task among CLIP models regardless of applied corruptions. The performance of PathCLIP is more steady than other two CLIP models when the severity level changes. PLIP ranks second in pathology image classification, providing evidence that training a model using specialized domain data can yield better performance than the OpenAI-CLIP model trained on more general data. Moreover, the performance of PLIP is largely affected by brightness, contrast, Guassian blur, saturation, and hue. It might be useful to include these image corruptions during training to improve model performance.

In image-image retrieval, PLIP surpasses the other two CLIP models. This might be attributed to the fact that the training images of PLIP are in high-resolution, whereas PathCLIP uses low resolution figures of research papers acquired from Pubmed. This discrepancy may result in inferior performance for PathCLIP compared to PLIP. In addition, the performance of OpenAI-CLIP can be influenced by the corruption related to brightness, blur and hue in the top 5 retrieved images. Nevertheless, the performance of PathCLIP and PLIP remains relatively competitive on image-image retrieval. 

\section{Conclusion}
In this work, we evaluate the robustness of PathCLIP for pathology image analysis. Specifically, we investigate the performance of PathCLIP on seven common types of corruptions in the real world. Our findings indicate that PathCLIP is relatively robust to corruptions and outperforms OpenAI-CLIP and PLIP in zero-shot classification. Among the seven corruptions, blur and resolution can significant affect the performance of the PathCLIP. Therefore, it is important to ensure image quality before applying a clinical test. We also assess the robustness of PathCLIP in the task of image-image retrieval. Results show that PathCLIP exhibits inferior performance to PLIP on Osteosarcoma under various corruptions. Experimental findings suggest that PLIP might be more suitable for pathology image analysis of bone cancer. For the purpose of clinical use, it is recommended to flexibly use one of the CLIP models depending on the tasks. Overall, PathCLIP shows promise as a foundational model for zero-shot classification and image-image retrieval in pathology images. Future work will consider image corruptions during model training to achieve robust performance. Another direction will focus on the development of PathCLIP and a large language model to perform deep multimodal understanding where AI can comprehend and generate responses or pathological diagnoses based on both textual and visual inputs.

\bmhead{Acknowledgments}
This work was supported in part by grants from the National Natural Science Foundation of China (Grant No. 92270108, No. 282302180, No. 82302180), Chinese National Key Research and Development Project (Grant No. 2021YFC2500400 and Grant No.2021YFC2500402) and Tianjin Key Medical Discipline (Specialty) Construction Project (TJYXZDXK-010A).

\bibliography{sn-bibliography}% common bib file

\end{document}